\documentclass[journal]{IEEEtran}

\usepackage{orcidlink}
\usepackage{graphicx}
\usepackage{amsmath, amssymb}
\usepackage{longtable}
\usepackage{tabularx}       
\usepackage{array}
\newcolumntype{P}[1]{>{\raggedright\arraybackslash}p{#1}}
\usepackage{authblk}
\usepackage{algorithm}
\usepackage{algpseudocode}
\usepackage{xcolor}
\usepackage{tikz} 
\usetikzlibrary{positioning, arrows.meta, shapes.geometric}
\usepackage[a4paper, margin=1in]{geometry}
\usepackage{hyperref}
\usepackage{float} 
\usepackage{multirow}
\usepackage{array}  

\begin{document}
\title{Reading Legends on Ancient Coins: An Object Detection Approach for Character Recognition on a Novel Roman Republican Dataset}

\author{%
    Hafeez Anwar \orcidlink{0000-0001-9529-3966}
    \thanks{Hafeez Anwar is with the Department of Computer Science, National University of Computer and Emerging Sciences (FAST-NUCES), Peshawar, 25000, Pakistan (email: hafeez.anwar@nu.edu.pk).} 
    
}
\markboth{}%
{Hafeez \MakeLowercase{\textit{et al.}}: Legends Recognition on Ancient Coins}

\maketitle
\begin{abstract}
When it comes to the proper classification of ancient coins with respect to their time and issuer, the textual inscriptions on these coins, also known as legends, are of paramount importance. These legends consist of alphabets/characters still used in English. This paper addresses image-based character recognition on ancient Roman Republican coins via a deep learning–based object detection strategy. However, legends on these coins pose high variation due to non-uniform placement, primitive inscription techniques, and wear and tear. Additional challenges include inconsistent imaging conditions such as illumination, orientation, and scale. To accommodate these, we gathered a novel large-scale dataset of 5,654 Roman Republican coin images, manually annotated with 21 character labels, totaling 38,808 annotations. For recognition, we use You Only Look Once (YOLO) variants: YOLOv3, v4, v5, v7, and v8. YOLOv7-Large achieves the best mAP50 of 90.4\%, followed by YOLOv7-Extended and YOLOv7-xl with 90.2\% and 90.1\%, respectively.
\end{abstract}
\IEEEpeerreviewmaketitle

\section{Introduction}
Image-based object detection is a subfield computer vision that deals automatically classifying images or their parts into predefined classes based on their depicted contents~\cite{torralba2024foundations}. This is a foundational step for a multitude of computer vision applications where the central task is to recognize objects of interest within images and assign them to their predefined classes or categories. The recent advances in the field of deep learning~\cite{torralba2024foundations} has also proved instrumental in the evolution of image-based object detection. Consequently, more advanced algorithms and frameworks are being proposed that have achieved remarkable performance on the tasks object detection and recognition.

\begin{figure*}[t!]
    \centering
    \includegraphics[width=1\textwidth]{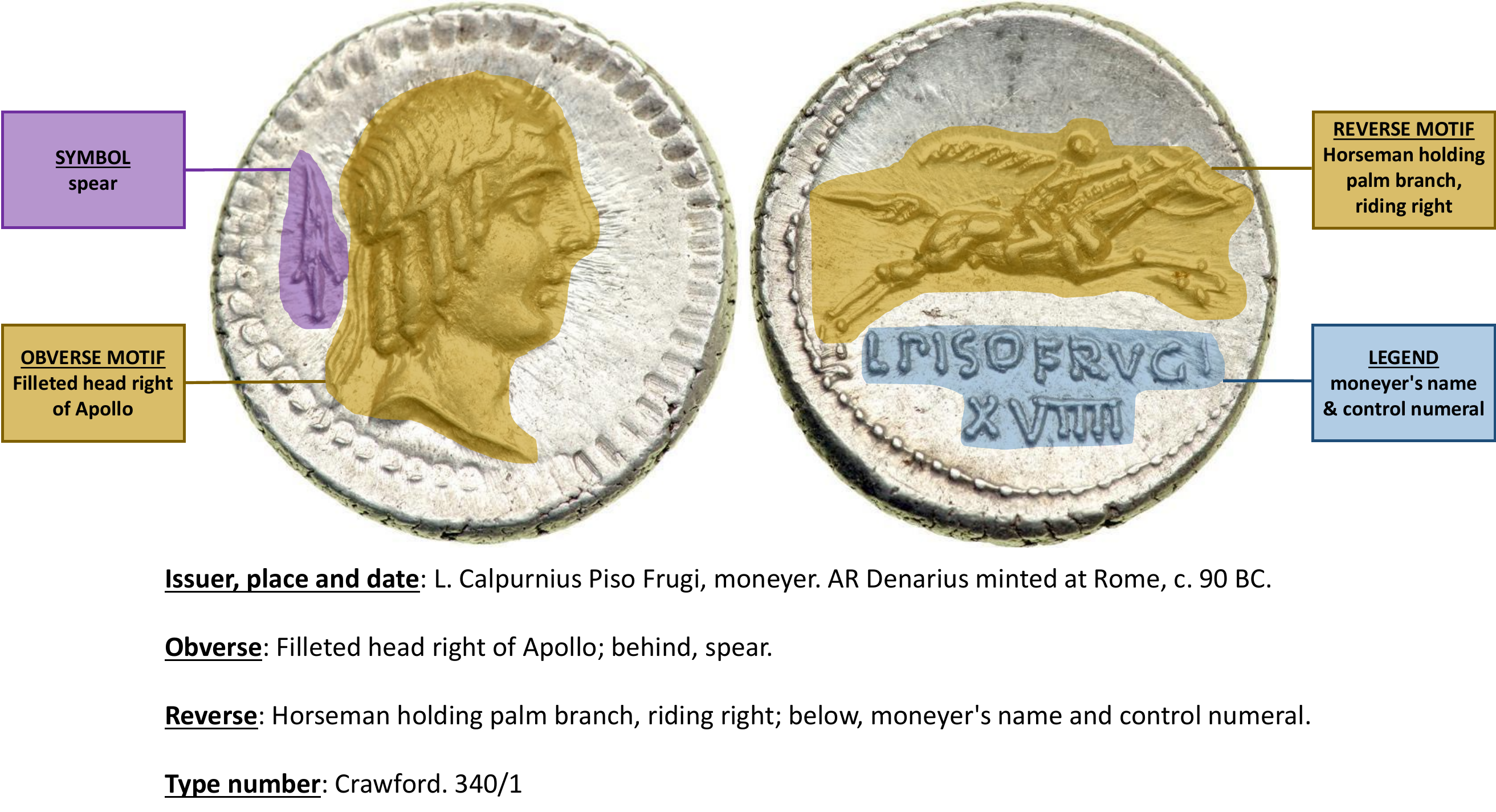}
    \caption{Various parts of a Roman Republican coin and how they are described textually in standard reference books. The legends are words that describe various facts; for instance, here, the name of the moneyer is shown along with the control mark that was used to avoid forgery. This coin is indexed by Crawford~\cite{crawford1974roman} given the type number as Crawford 340/1}
    \label{fig:1}
\end{figure*}

When it comes to the application of computer vision methods, the relatively recent field of Digital Humanities is yet to be discovered. To this end, the image-based object detection posses a transformative potential. The vast image collections owned both by individual and the museums require extensive efforts to get properly managed and organized. This can be achieved via the object detection methods applied to visual data of art history~\cite{madhu2020understanding}, archeology~\cite{anwar2014rotation}, and other related disciplines. For instance, in art history, object detection methods can be used to recognize artworks~\cite{bernasconi2023computational} based on style, period, or artist, thus enabling scholars to uncover patterns and connections that might otherwise remain hidden. In archeology, it can assist in the automated recognition of artifacts, helping researchers to more efficiently analyze, arrange and interpret archaeological findings. Furthermore, by applying image-based object detection to historical photographs or documents, historians can systematically categorize visual materials that results in streamlining the process of archiving and preserving cultural heritage~\cite{muehlberger2019transforming}. This interdisciplinary application of image-based object detection not only enhances the efficiency of research in digital humanities but also opens new avenues for exploring and understanding the visual aspects of human history and culture.

This paper deal with the image-based analysis of the ancient Roman Republican coins. This is a motivating example where object detection and image classification can be used to categorize coins based on various attributes such as design, era, and region of origin. While in the past, coins were used for trade and monetary purposes, today, ancient coins are considered a rich source of information about historical events, personalities and places. In addition to that, these coins are considered precious antiques and historical artifacts. Consequently, the ancient coins trade is an ever-flourishing business that involves collectors, enthusiasts, investors, historians and even museums. The peculiar appearance of these coins reveals the expertise of the engravers of the particular time to which they belong. Similarly, the engraved objects, text and monuments etc., disclose the cultural norms and values of the habitants who used these coins for various purposes. We, in this paper, are interested in the ancient coins from the Roman Republican era that are categorized by Crawford~\cite{crawford1974roman} into ``coin types'' based on various metrics such as era, engraver and the place or authority of issuance. On the majority of coin types, the obverse side depicts a portrait such as of Roma, and the reverse side depicts various places, objects, animals, soldiers and riders etc. The anatomy of a typical Roman Republican coin and its textual description is shown in Figure~\ref{fig:1}. The textual descriptions of these parts, the information about the issuer and the issuance dates are mentioned in the reference books under each coin type. 

\begin{figure*}[t!]
    \centering
    \includegraphics[width=\textwidth]{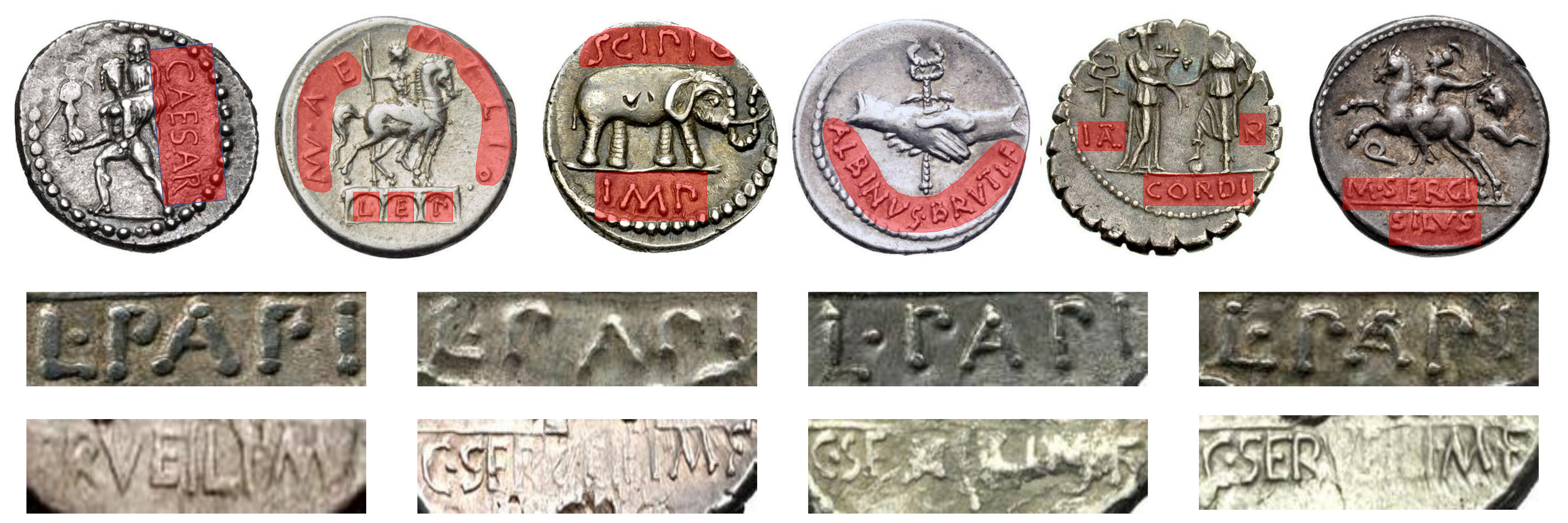}
    \caption{Visual challenges due to non-uniformity of legends. The first row shows the non-uniform position of the legends. The second row shows the same legends but with different appearances due to non-uniform dies. The third row shows the non-uniformity induced by wear and tear.}
    \label{fig:2}
\end{figure*}

Among others, the legends, i.e., the alphabets and words on ancient coins, carry important historical information. More notably, it used to be a source of political propaganda that transformed into several stages throughout the Republican era~\cite{luce1968political}. To summarize the coin legends reveal important archaeological information that is considered vital in understanding historical events. Due to this reason, this paper deals with image-based recognition of characters which is considered as a first step in the automatic recognition of legends in the coin images.

Unlike the modern-day coins, the legends engraved on ancient coins are highly non-uniform due to a number of factors. 
First, the manufacturing process of the ancient coins is completely different from the modern day high precision mechanized minting process. Ancient coins were manufactured either manually or with imprecise die casts due to which the exact replication of coins became highly impractical. Due to this reason, in addition to other coin parts, a high degree of variations could be found in the shape and placement of the coins legends. Consequently, the legends inscribed on coins of the same type appear differently due to the non-uniformity found in the visual appearance of their individual letters. Apart from the imprecise dies casts and the manual handicraft of the engravers, in order to achieve bulk production, the geological minting locations of coins~\cite{crawford1974roman} were also different thus causing even more variations in the legends appearance. Consequently, all these factors led to a non-uniform arrangement of the legends across different coin specimens of the same type.

Second, the prolonged age of ancient coins further intensifies the non-uniformity found in their legends.  
Ancient coins suffer from an immense wear and tear caused by the natural corrosion process as most of the times they are buried in soil at archaeological sites. In addition to wear and tear, the corrosion process also adds an unwanted layer of minerals and chemicals on the parts of the coin surface. Coin legends, in particular fall prey to such layer formations as they have empty spaces between their individual letters. Such factors i.e. prolonged age and hostile preserving conditions lead to the visual deterioration of coin legends thus making them unreadable and difficult to recognize. As a result, legends of the same coin type appear differently on different coin specimens. 

Lastly, in case of the Roman Republican coins, there is no standardized system to decide the place of legend on the coin surface thus adding yet another cause of variability found in the legends. Unlike the Roman Imperial coins~\cite{arandjelovic2012reading} that follow a uniform standard to place the legend along the coin border, the Roman Republican coins do not follow such uniformity to fix the place of legends. Due to such non-uniformity, the legends on the Roman Republican coins could be found at various positions both horizontally and vertically, at various orientations, along the border of the coin, and even combined with the main motif of the coin.

Overall, the non-uniformity of legends on ancient coins is the result of a combination of factors, including the imprecise manufacturing techniques of the time, the effects of aging and environmental damage, and the absence of standardized practices for legend placement. This inherent variability presents both challenges and opportunities for historians, numismatists, and researchers who seek to decode and understand the historical context and significance of these ancient artifacts.

Figure~\ref{fig:2} depicts some exemplar images of Roman Republican coins representing all these challenges. The variations described so far are seen as one set of challenges in the image-based recognition of the legends of alphabets posed by the ancient coins themselves. A second set of challenges is due to the variations caused by the imaging condition of the coins. For instance, coins of the same type and almost similar appearance can appear visually dissimilar due to the non-uniform illumination~\cite{zambanini2014classifying} used for its imaging. Similarly, coins being circular objects, can exhibit in-plane orientations that can cause intra-class variations among coins of similar types. Other potential variations are differences in coin scale and position~\cite{anwar2014rotation}.    
\begin{figure*}[t!]
    \centering
    \includegraphics[width=\textwidth]{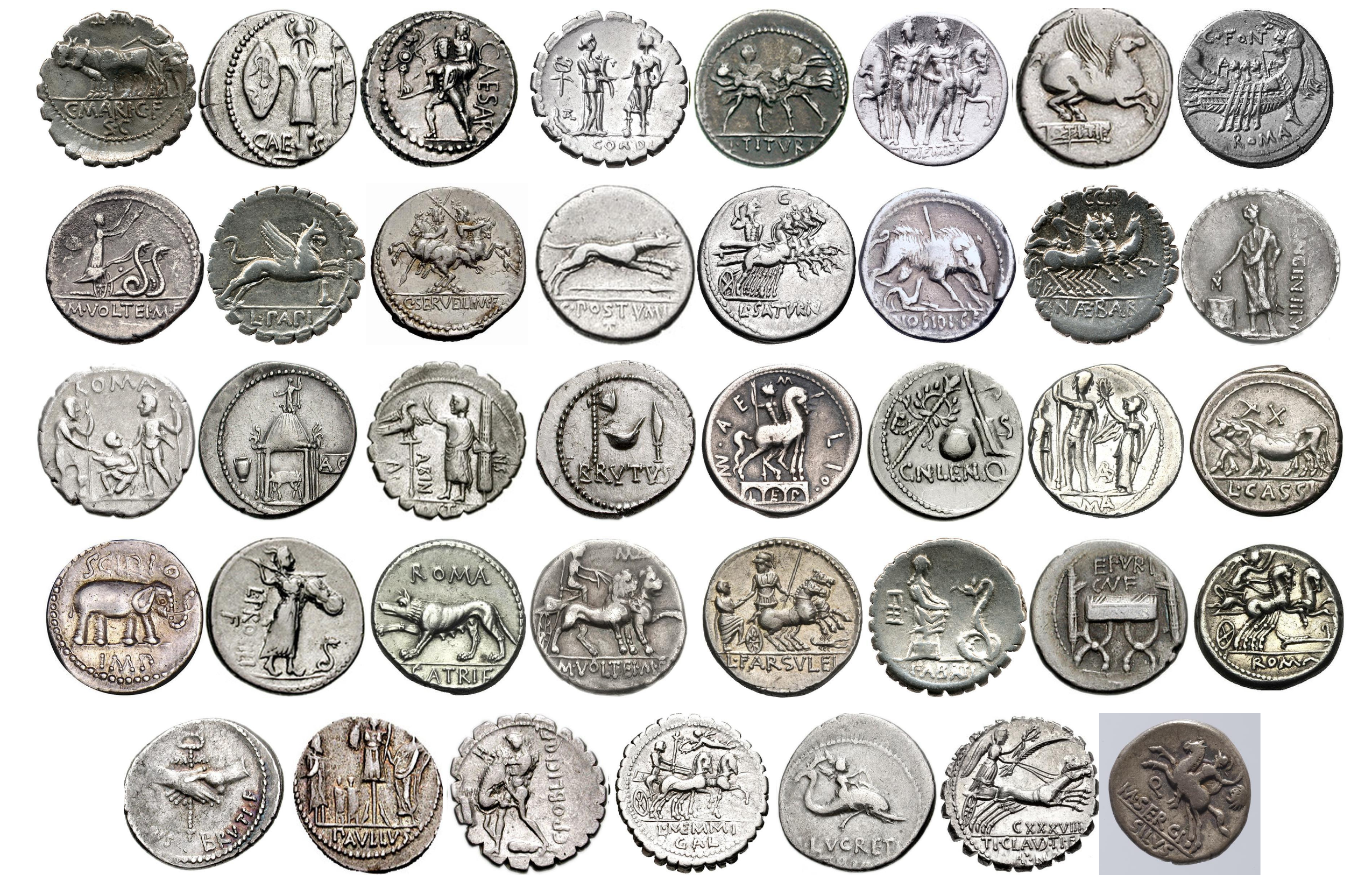}
    \caption{Exemplar images of 39 coin types included in the dataset. It is clearly observable that the legends positions on the selected coin type vary greatly making our curated dataset very challenging for the deep learning-based object detection frameworks}
    \label{fig:3}
\end{figure*}

\section{Literature review}
Despite their importance in ancient coin classification, the image-based recognition of legends on ancient coins has drawn less attention compared to obverse side portrait recognition~\cite{schlag2017ancient} and reverse side motif recognition~\cite{anwar2014rotation}. This is mainly related to the fact that on ancient coins, legends are relatively more affected by wear and tear than other parts of the coins. However, despite this weakness, they present themselves as rich source of vital information, making their image-based recognition non-trivial and appealing to the Numismatics community. 

The legend recognition on the Roman Republican coins was first done by Kavelar et al.~\cite{kavelar2012word}, who also approached the problem using an object recognition paradigm. 
The paper presents a novel approach for recognizing legends, or textual inscriptions, on ancient Roman coins using advanced computer vision techniques. Traditional Optical Character Recognition (OCR) methods are unsuitable for this task due to the unique challenges posed by the curved, degraded surfaces of the coins and the lack of clear character segmentation. Instead, the authors propose using individual character classifiers applied to a dense grid of local SIFT features~\cite{lowe1999object}. The final word recognition is accomplished through a lexicon-based method using the Pictorial Structures approach. The method was tested on 180 images of Roman coins with 35 different legend words, achieving word detection rates between 29\% and 53\%, depending on the lexicon size. This research highlights the potential of object recognition techniques in challenging real-world scenarios where conventional OCR fails.
This method achieved better recognition rates when combined with the local features matching~\cite{zambanini2013improving} where  a novel method for classifying ancient Roman Republican coins using image-based techniques. The authors combine exemplar-based classification, which evaluates the visual similarity of coins through dense correspondence fields, with lexicon-based legend recognition, which identifies coin legends. By integrating classification scores from both the obverse and reverse sides of the coins, the proposed method enhances the accuracy of classification. Experiments conducted on a dataset of 464 coin images across 60 different classes demonstrate that the fusion of these methods results in a higher classification rate than using either approach independently hence contributing to the field of numismatics by providing a more robust and reliable system for automated coin classification.
Ancient Roman Imperial coins are classified via legend recognition\cite{arandjelovic2012reading}. The legend is assumed to be located along the coin border and is curvature normalized by a log-polar transformation. However, this assumption does not hold for the Roman Republican coins, as their legends' positions are not fixed. 
\begin{figure*}[t!]
    \centering
    \includegraphics[width=\textwidth]{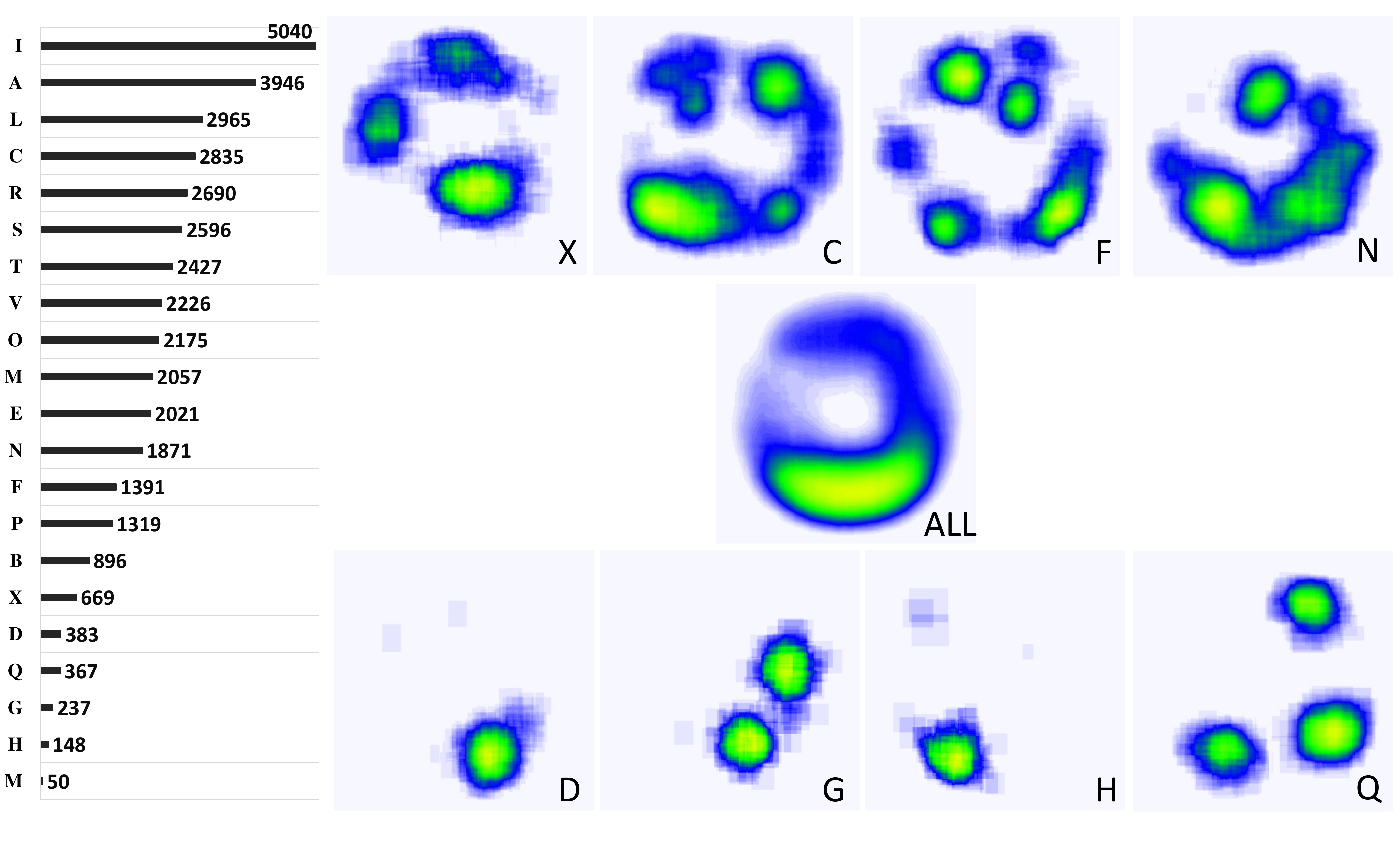}
    \caption{This figure shows the total number of character samples annotated across the entire coin image dataset. The heatmap illustrates the positions where these characters appear on the coins, based on the annotations. Brighter areas indicate positions with higher character density. This visualization helps us understand both the frequency and spatial layout of characters, which is important for training and evaluating character detection models. The heatmap clearly shows that the positions of some of the characters, such as ``X'', ``C'', ``F'', and ``N'' shown in the top row, vary greatly, while others, such as those shown in the bottom row; ``D'', ``G'', ``H'', and ``Q'', show little variation in their spatial positions on the coins. The middle row shows the overall heatmap of the characters' spatial occurrence, clearly indicating that most characters appear at the bottom of the coins, beneath the main motif.
}
    \label{fig:4}
\end{figure*}

We perform character recognition on Roman Republican coins, the first step for legends recognition. To this end, we treat each character as an object and perform its recognition via localization. Consequently, the following are our novel contributions,
\begin{enumerate}
    \item We gather the largest Roman Republican dataset for character recognition with 5,654 images.
    \item These images are manually annotated for 21 characters, where the total number of annotations is 38,808
    \item We evaluate several state-of-the-art deep learning-based object localization frameworks for individual character recognition, which are YOLOv5~\cite{jocher2020yolov5}, YOLOv7~\cite{wang2023yolov7}, and YOLOv8\cite{Jocher_Ultralytics_YOLO_2023}.
\end{enumerate}
\section{Dataset}
The process of preparing the image dataset spans two major steps namely; image collection and ground truth generation via manual annotations. We elaborate on these two steps in the following. 
\subsection{Images of the Roman Republican coins}
Our dataset consists of 5654 images that are collected from various sources both online and from various museums. As a result, images of the same coin types vary greatly not only due to the coins themselves but also due to the non-coherent imaging conditions. For instance, on the auctions websites, the coins are made attractive by applying special chemicals in order to increase their shine. The Roman Republican coins are indexed by Crawford~\cite{crawford1974roman} in more than 500 types. Nonetheless, we use a subset of those types and hence the images in our dataset belong to 39 types whose representative exemplar images are shown in Figure~\ref{fig:3}. It is also worth mentioning that, we use images of the reverse sides that differ from one another based on the main objects along with the legends. 
\subsection{Image annotation}
Since, it is the problem of localization, the ground truth is generated via creating bounding boxes around the characters of the legends. For this purpose, the manual annotation is done using Labelimg~\cite{labelimg} image annotation tool. The total number of annotated characters is 21 where the number of annotations per character are shown in Figure~\ref{fig:4}. The total number of annotations are 38,808. The hit-map of some of the characters are also shown where the first row shows the highly scattered while the bottom row shows the least scattered characters. The middle row shows the cumulative hit-map of all the characters. Hence, the complication can be observed that unlike the Roman Imperial coins~\cite{arandjelovic2012reading}, the legends of the Roman Republican coins are highly non-uniform with respect to their locations on the coins. 

\section{Methodology}
YOLO is the first ever object detector that unifies features extraction, object localization and classification in a single end-to-end network. This strategy makes YOLO completely different from detectors such as FasterRCNN that leverage upon a two step detection strategy where in the first step the region for an object presence is detected followed by the classification of the detected object. Such single pass object detection and classification results in better speed, scalability and usability[Redmon thesis] that is reflected in the later YOLO-based models. 

YOLOv1~\cite{redmon2016you} was the first single stage detector. The architecture had 24 convolutional and 4 maxpooling layers that are followed by two fully connected layers. This architecture utilized Batch Normalization for model regularization and leaky ReLU acitivations except the last layer where a linear activation function is used.  A smaller and faster version of YOLOv1 namely YOLO Fast was also proposed having 9 convolutional layers that contained less filters as compared to the full version.

YOLOv2~\cite{redmon2016you} was proposed with some changes done to YOLOv1 such as the removal of fully connected layers thus making the architecture independent of the input image resolution. The architecture contained 22 convolution and 5 maxpool operations while the output is a 125 dimensional feature vector  A smaller, faster and comparatively less accurate version of YOLOv2 namely Tiny-YOLOv2 was also proposed. 

Almost every YOLO-inspired architecture consists of a backbone, neck and head. YOLOv3~\cite{redmon2018yolov3} uses the Darknet-53 feature extraction network as a backbone, which consists of convolutional layers. Feature Pyramid Network (FPN) is introduced in YOLOv3, which allows object detection at various scales. 

In YOLOv4~\cite{bochkovskiy2020yolov4}, CSPDarknet53 is used as the backbone to extract features. Additional modules such as Spatial Pyramid Pooling (SPP) and PANet are employed in YOLOv4 as neck, which concatenates the features extracted from the input with the features extracted from the end layers. YOLOv3 dense prediction block is used in YOLOv4 as the head. 

The backbone in YOLOv5~\cite{jocher2020yolov5} consists of convolutional layers which encode input to the features map. The neck includes feature enhancers like PANet and SPP, improving detection accuracy and speed. The head part generates predictions by applying convolutional operations on the enhanced features. 

YOLOv7~\cite{wang2023yolov7} was another step forward in refining the balance between speed and accuracy. This version introduced new network designs and optimization techniques that further enhanced the model's performance in real-time applications. YOLOv7 focused on improving the efficiency of computations, making it possible to achieve higher accuracy with less computational overhead, thereby expanding its applicability to more resource-constrained devices.

YOLOv8\cite{Jocher_Ultralytics_YOLO_2023} represents the latest in the evolution of the YOLO series, incorporating all the lessons learned from previous versions. It introduces even more advanced techniques for feature extraction, better optimization strategies, and further improvements in handling small objects and complex scenes. YOLOv8 is designed to be the most robust and versatile version yet, capable of achieving high accuracy with efficient computation, making it suitable for a wide range of applications from cloud-based systems to edge devices.

\section{Results and discussion}
We use a pre-trained model of each YOLO variant and then fine-tune it on our Roman Republican coins dataset. These models are trained on Intel Core i9-11900 11$^{th}$ Generation processor, NVIDIA RTX 3050 graphics card and 32GB RAM. The models were evaluated based on their precision, recall, mean average precision (mAP) at 50\% IoU (Intersection over Union), and mAP from 50\% to 95\% IoU, providing a comprehensive overview of their performance across different levels of detection thresholds. 

\begin{table}[t!]
\centering
\caption{Performance metrics of all the utilized YOLO variants. For YOLOv3 and YOLOv4 2000 epochs are performed to achieve results comparable with the rest of the other variants. For all the variants of YOLOv5, YOLOv7 and YOLOv8 100 epochs are used. }
\label{tab:1}
\begin{tabular}{lcccc}
\hline
\hline
Model       & Precision & Recall & mAP50 & mAP50-95 \\ \hline
YOLOv3		  & 64				& 81		 & 78.5	 &  49.35		\\
YOLOv3-Tiny	& 78				& 77		 & 75.6	 &  59.77		\\ \hline
YOLOv4      & 80				&	93		 & 88.5	 &  61.7		\\
YOLOv4-Tiny & 86				&	89		 & 85.9	 &  69.14		\\ \hline
YOLOv5-s    & 87.6      & 85.4   & 86.4  &  59.7     \\
YOLOv5-m    & 88        & 86.8   & 87.2  &  60.8     \\
YOLOv5-l    & 87.7      & 87.2   & 86.9  & 60.8     \\
YOLOv5-n    & 88        & 82.1   & 85.6  & 56.7     \\
YOLOv5-x    & 88.7      & 87.1   & 88.1  & 60.7     \\ \hline
YOLOv7-l    & 87.9      & 86.7   & 90.4  & 63.3     \\
YOLOv7-tiny & 87.5      & 84.9   & 87.7  & 58       \\
YOLOv7-x    & 87.6      & 89.6   & 90.1  & 62.7     \\
YOLOv7-D6   & 88.6      & 86.2   & 89.8  & 62.4     \\
YOLOv7-E6   & 88.3      & 85.4   & 88.5  & 61       \\
YOLOv7-E6E  & 86.8      & 86.7   & 90.2  & 62.2     \\
YOLOv7-W6   & 89.1      & 84.7   & 88.4  & 61.5     \\ \hline
YOLOv8-s    & 86.9      & 87.1   & 88.1  & 62.6     \\
YOLOv8-m    & 87.2      & 87.2   & 88    & 63       \\
YOLOv8-l    & 89.2      & 86.7   & 88.9  & 63.1     \\
YOLOv8-n    & 88.8      & 84.3   & 88.5  & 62.1     \\
YOLOv8-x    & 87.5      & 87.4   & 89.1  & 63.3     \\ \hline
\hline
\end{tabular}%
\end{table}

\begin{figure*}[t!]
    \centering
    \includegraphics[width=0.9\textwidth]{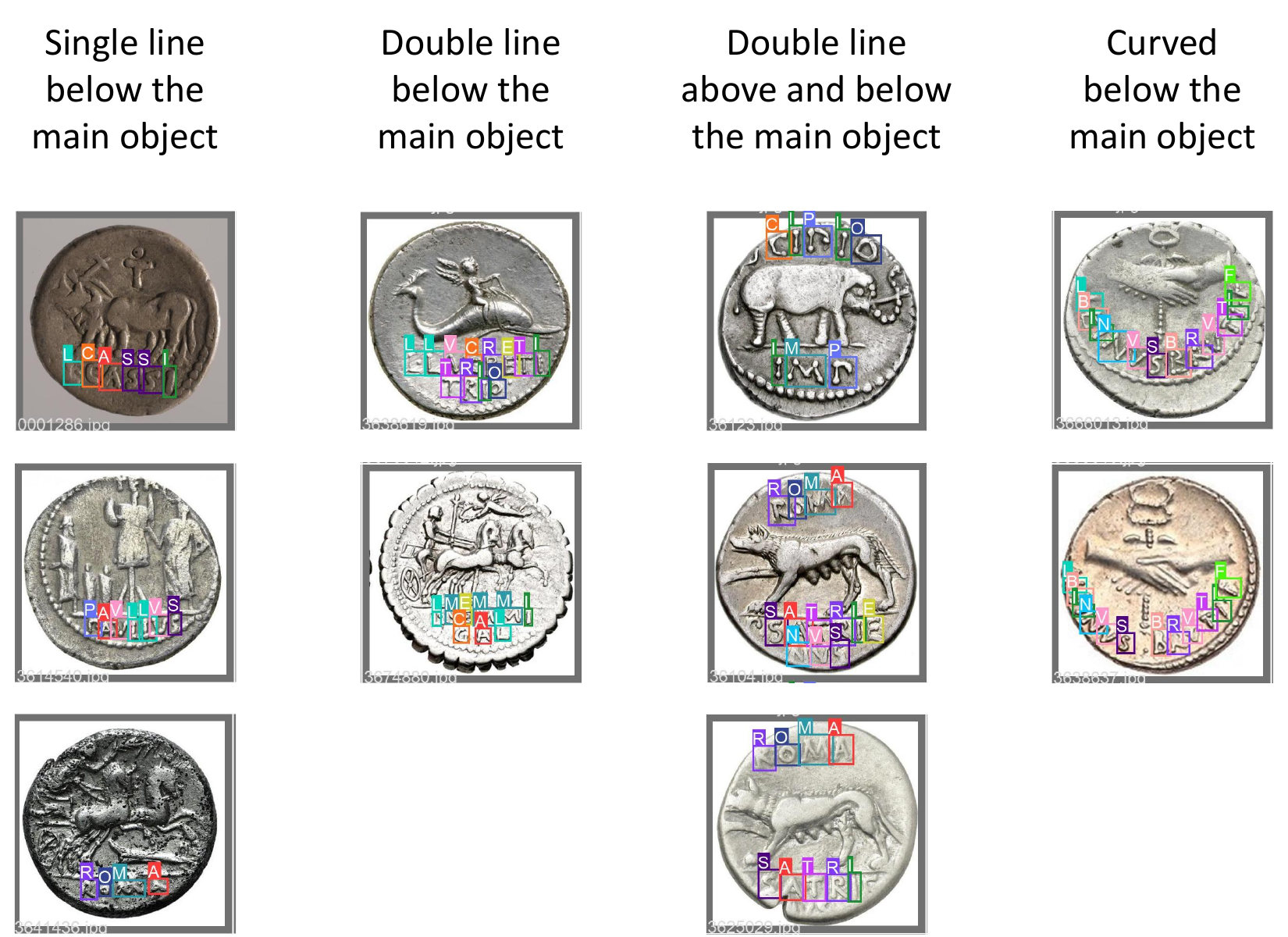}
    \caption{Characters detection results on coins with various legends including horizontal, curved, multiple legends above and below the main object and vertical legends}
    \label{fig:5}
\end{figure*}
We utilized a variety of YOLO (You Only Look Once) architectures to tackle the challenge of character recognition on Roman Republican coins. The legends of these coins present unique difficulties for the task of object detection due to their irregular character placements, varying states of preservation, and the non-standardized imagery resulting from diverse historical minting practices. The objective is to determine which YOLO variant could most effectively perform image-based recognition of the characters inscribed on these ancient artifacts.
Table~\ref{tab:1} shows the comprehensive results where YOLOv7 achieves the best performance. The detection results are in Figure~\ref{fig:5}. The table presents a comparative analysis of various models from the YOLO (You Only Look Once) family, including versions 3, 4, 5, 7, and 8, across different scales (s, m, l, n, x). The models are evaluated based on four key performance metrics: Precision, Recall, mAP\@50, and mAP\@50-95. Following is the detailed discussion of the achieved results:

\subsection{Precision}
Precision measures the accuracy of positive predictions, indicating how often the model's predicted objects were correctly identified. Among all the evaluated models, YOLOv8-l achieved the best precision value of 89.2\% followed closely by YOLOv7-W6 at 89.1\%. This demonstrates the effectiveness of these models at reducing the false positives, thus making them reliable in situations that require the correct detection of objects. Other models such as YOLOv8-n and YOLOv5-x also achieved nearly strong precision thus making them competitive options.

\subsection{Recall}
Recall is the measurement of a model's ability to detect all the relevant objects which means that what percentage of actual objects are recognized by the model. Among the evaluated models, YOLOv7-x achieved the highest recall rate of 89.6\% which demonstrates its capability of detecting most of the characters in the dataset. Such high recall could prove crucial for future extension of the current work such as the recognition of complete legends which requires the detection of all the characters in a given legend. YOLOv5-l and YOLOv8-s follow closely where both the models achieve a recall rate of 87.2\% thus making them both the second best choice for relevant characters detections. 
\begin{figure*}[t!]
    \centering
    \includegraphics[width=1\textwidth]{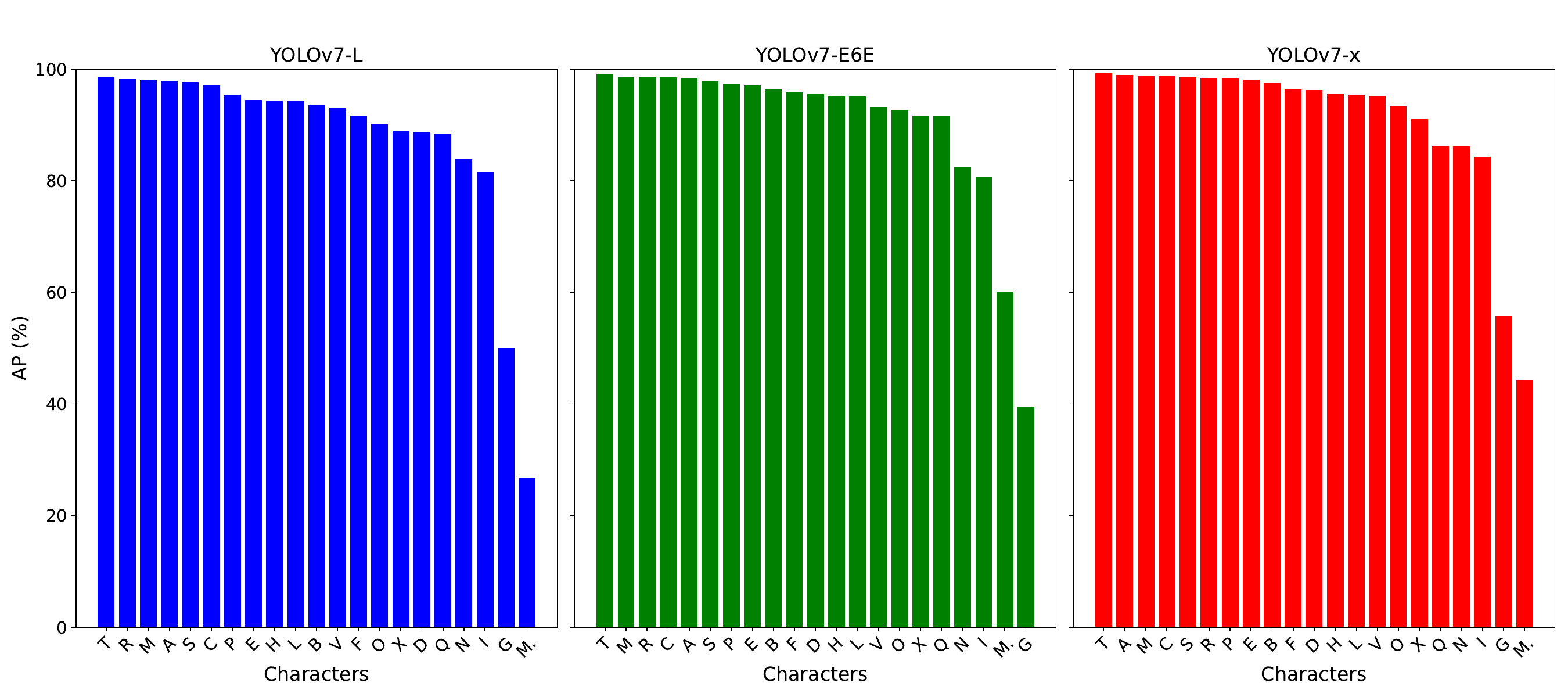}
    \caption{Comparison of character-wise Average Precision (AP) scores achieved by the Top-3 three YOLOv7 variants - YOLOv7-L, YOLOv7-E6E, and YOLOv7-X, on the ancient coin dataset. Each bar represents the AP score for a specific Roman character. While all models show good performance on frequently occurring characters like ``T'', ``A'', and ``M'', there are notable differences in how they handle less frequent or visually similar characters such as ``Q'', ``G'', and ``I''. This visualization helps highlight which characters are more challenging for the models to detect reliably.}
    \label{fig:6}
\end{figure*}

\subsection{mAP50 (Mean Average Precision at IoU 50\%)} 
mAP50 is the mean of average precisions achieved over all the classes. For mAP50, the threshold value for Intersection over Union (IoU) is taken as 50\%. This is the most commonly used metric to evaluate the overall performance of object detection models. 
With respect to mAP50, YOLOv7-l outperforms the rest of the models by achieving a score of 90.4\% demonstrating its strong ability on the task of characters detection in the images of ancient Roman Republican coins. YOLOv7-E6E follows closely with a score of 90.2\% that is succeeded by YOLOv7-x achieving a mAP50 value of 90.1\%.

\subsection{mAP50-95 (Mean Average Precision across IoU thresholds from 50\% to 95\%)}
This is an even more challenging metric that gives the mean of average precisions calculated over multiple Intersection over Union (IoU) thresholds. 
Among the evaluated models, surprisingly, YOLOv4 attains the best mAP50-95 value which is 69.1\% followed by YOLOv8-x and YOLOv7-l both achieving the same score of 63.3\%. 

This metric indicates that these models are not only accurate but also versatile in handling varying levels of overlap between detected objects and their true positions.

\subsection{Average Precision (AP)}
Figure~\ref{fig:6} shows the average precision achieved by the top 3 performing models at each character class which are YOLOv7-L, YOLOv7-E6E, and YOLOv7-x. It can be observed that all the three models perform exceptionally well for characters A, C, M, S, and T where they achive AP of more than 98\% while for characters other than M. and G, they all perform well above the 85\% mark. The lowest APs for the two characters can be attributed to the fact that they have far less images than the rest of the characters as can be seen in Figure~\ref{fig:4}. However, this may not be the sole reason for such a low AP as the number of images for character H are less than those of character G. This is attributed to yet another variation that we have already discussed and that is the one caused by the variations in placement. It can be observed in Figure~\ref{fig:4} that the place of character H is fixed while character G could be found on three different places, and thus we believe that is a strong reason that the value of AP of G is well belove than that of H. 

\subsection{Overall Analysis}
YOLOv7 models generally show a strong performance across all metrics, particularly in mAP50 and mAP50-95, making them highly effective for precise and consistent object detection.
YOLOv8 models, particularly YOLOv8-x and YOLOv8-l, also perform exceptionally well, especially in terms of precision and mAP50-95, highlighting their ability to accurately detect objects with minimal errors.
YOLOv5 models, while slightly lower in some metrics compared to the newer versions, still demonstrate solid performance, particularly YOLOv5-x, which balances high precision with good recall and mAP scores.
\subsection{Concluding the achieved performances}
The table suggests that while all models offer strong performance, the YOLOv7 and YOLOv8 models, particularly in their larger configurations (like -x and -l), provide the best overall balance of precision, recall, and mAP scores. These models are likely the best candidates for applications requiring high accuracy in object detection across various levels of overlap and complexity. The choice of the best model may depend on the specific application needs, but YOLOv7 and YOLOv8 series models stand out as the most capable overall.

\subsection{Future Directions:}
Continued improvements in model architecture could further enhance accuracy and efficiency. Additionally, expanding the dataset to include more varied coin types and further tuning the models to account for extreme cases of wear and damage could generalize the models' applicability to broader numismatic or archaeological applications. Exploring other deep learning frameworks and comparing their effectiveness against the YOLO variants could also provide new insights and possibly better solutions to the challenges involved in the image-based characters recognition on ancient Roman Republican coins.

\section{Conclusions}
We perform character recognition on ancient Roman Republican coins using an object detection paradigm where each character is considered as an object to be recognized via detection. This is motivated by the fact that the legends on Roman Republican coins vary significantly due to their positions on the coins. Other challenges include visual degradation caused by wear and tear and the variation due to imaging, such as non-uniform illuminations. We gather a novel dataset of 5654 images belonging to 39 coin types that are annotated manually for 21 characters making the total number of annotations 38,808. We evaluated variants of the YOLO object detector for character recognition on Roman Republican coins, namely YOLOv3, YOLOv4, YOLOv5, YOLOv7, and YOLOv8 where a maximum mAP(\@0.5) of 91.4\% is achieved with YOLOv7-l. 

This study shows that deep learning-based object detectors can be effectively used for reading legends on historical artifacts like coins, despite the challenging nature of the data. Our dataset and findings can serve as a foundation for future research in the field of digital numismatics, especially for tasks involving automated cataloging, search, and preservation. We hope that this work encourages further exploration of machine learning techniques for analyzing cultural heritage datasets.

\bibliographystyle{ieeetr}
\bibliography{egbib}

@book{torralba2024foundations,
  title={Foundations of computer vision},
  author={Torralba, Antonio and Isola, Phillip and Freeman, William T},
  year={2024},
  publisher={MIT Press}
}

@article{muehlberger2019transforming,
  title={Transforming scholarship in the archives through handwritten text recognition: Transkribus as a case study},
  author={Muehlberger, Guenter and Seaward, Louise and Terras, Melissa and Oliveira, Sofia Ares and Bosch, Vicente and Bryan, Maximilian and Colutto, Sebastian and D{\'e}jean, Herv{\'e} and Diem, Markus and Fiel, Stefan and others},
  journal={Journal of documentation},
  volume={75},
  number={5},
  pages={954--976},
  year={2019},
  publisher={Emerald Publishing Limited}
}

@article{bernasconi2023computational,
  title={A Computational Approach to Hand Pose Recognition in Early Modern Paintings},
  author={Bernasconi, Valentine and Cetini{\'c}, Eva and Impett, Leonardo},
  journal={Journal of Imaging},
  volume={9},
  number={6},
  pages={120},
  year={2023},
  publisher={MDPI}
}

@inproceedings{redmon2016you,
  title={You only look once: Unified, real-time object detection},
  author={Redmon, Joseph and Divvala, Santosh and Girshick, Ross and Farhadi, Ali},
  booktitle={Proceedings of the IEEE conference on computer vision and pattern recognition},
  pages={779--788},
  year={2016}
}

@misc{Jocher_Ultralytics_YOLO_2023,
author = {Jocher, Glenn and Chaurasia, Ayush and Qiu, Jing},
license = {AGPL-3.0},
title = {{Ultralytics YOLO}},
url = {https://github.com/ultralytics/ultralytics},
version = {8.0.0},
year = {2023}
}

@inproceedings{wang2023yolov7,
  title={{YOLOv7}: Trainable bag-of-freebies sets new state-of-the-art for real-time object detectors},
  author={Wang, Chien-Yao and Bochkovskiy, Alexey and Liao, Hong-Yuan Mark},
  booktitle={Proceedings of the IEEE/CVF Conference on Computer Vision and Pattern Recognition (CVPR)},
  year={2023}
}

@inproceedings{madhu2020understanding,
  title={Understanding Compositional Structures in Art Historical Images Using Pose and Gaze Priors: Towards Scene Understanding in Digital Art History},
  author={Madhu, Prathmesh and Marquart, Tilman and Kosti, Ronak and Bell, Peter and Maier, Andreas and Christlein, Vincent},
  booktitle={European Conference on Computer Vision},
  pages={109--125},
  year={2020},
  organization={Springer}
}

@misc {labelimg,
    author = "tzutalin",
    title  = "LabelImg. Git code (2015)",
    year   = "2015",
    url    = "https://github.com/tzutalin/labelImg"
}

@article{jocher2020yolov5,
  title={yolov5},
  author={Jocher, Glenn and Nishimura, K and Mineeva, T and Vilari{\~n}o, R},
  journal={Code repository},
  year={2020}
}

@article{redmon2018yolov3,
  title={Yolov3: An incremental improvement},
  author={Redmon, Joseph and Farhadi, Ali},
  journal={arXiv preprint arXiv:1804.02767},
  year={2018}
}

@article{bochkovskiy2020yolov4,
  title={Yolov4: Optimal speed and accuracy of object detection},
  author={Bochkovskiy, Alexey and Wang, Chien-Yao and Liao, Hong-Yuan Mark},
  journal={arXiv preprint arXiv:2004.10934},
  year={2020}
}

@inproceedings{lowe1999object,
  title={Object recognition from local scale-invariant features.},
  author={Lowe, David G and others},
  booktitle={ICCV},
  volume={99},
  number={2},
  pages={1150--1157},
  year={1999}
}

@inproceedings{kavelar2012word,
  title={Word detection applied to images of ancient roman coins},
  author={Kavelar, Albert and Zambanini, Sebastian and Kampel, Martin},
  booktitle={International Conference on Virtual Systems and Multimedia},
  pages={577--580},
  year={2012},
}

@inproceedings{zambanini2013improving,
  title={Improving ancient roman coin classification by fusing exemplar-based classification and legend recognition},
  author={Zambanini, Sebastian and Kavelar, Albert and Kampel, Martin},
  booktitle={ICIAP},
  pages     = {149--158},
  year={2013}
}

@inproceedings{zambanini2014classifying,
  title={Classifying ancient coins by local feature matching and pairwise geometric consistency evaluation},
  author={Zambanini, Sebastian and Kavelar, Albert and Kampel, Martin},
  booktitle={ICPR},
  pages={3032--3037},
  year={2014},
}

@inproceedings{arandjelovic2012reading,
  title={Reading ancient coins: automatically identifying denarii using obverse legend seeded retrieval},
  author={Arandjelovi{\'c}, Ognjen},
  booktitle={ECCV},
  pages={317--330},
  year={2012},
}

@inproceedings{schlag2017ancient,
  title={Ancient Roman coin recognition in the wild using deep learning based recognition of artistically depicted face profiles},
  author={Schlag, Imanol and Arandjelovic, Ognjen},
  booktitle={ICCV Workshop)},
  pages = {2898-2906},
  year={2017},
}

@book{crawford1974roman,
  title={Roman republican coinage},
  author={Crawford, Michael H},
  volume={1},
  year={1974},
  publisher={Cambridge University Press}
}

@article{luce1968political,
  title={Political propaganda on Roman Republican coins: circa 92-82 BC},
  author={Luce, Torry J},
  journal={American journal of archaeology},
  volume={72},
  number={1},
  pages={25--39},
  year={1968},
  publisher={The University of Chicago Press}
}

@inproceedings{anwar2014rotation,
  title={A rotation-invariant bag of visual words model for symbols based ancient coin classification},
  author={Anwar, Hafeez and Zambanini, Sebastian and Kampel, Martin},
  booktitle={2014 IEEE International Conference on Image Processing (ICIP)},
  pages={5257--5261},
  year={2014},
  organization={IEEE}
}

\begin{IEEEbiography}[{\includegraphics[width=1in,height=1.25in,clip,keepaspectratio]{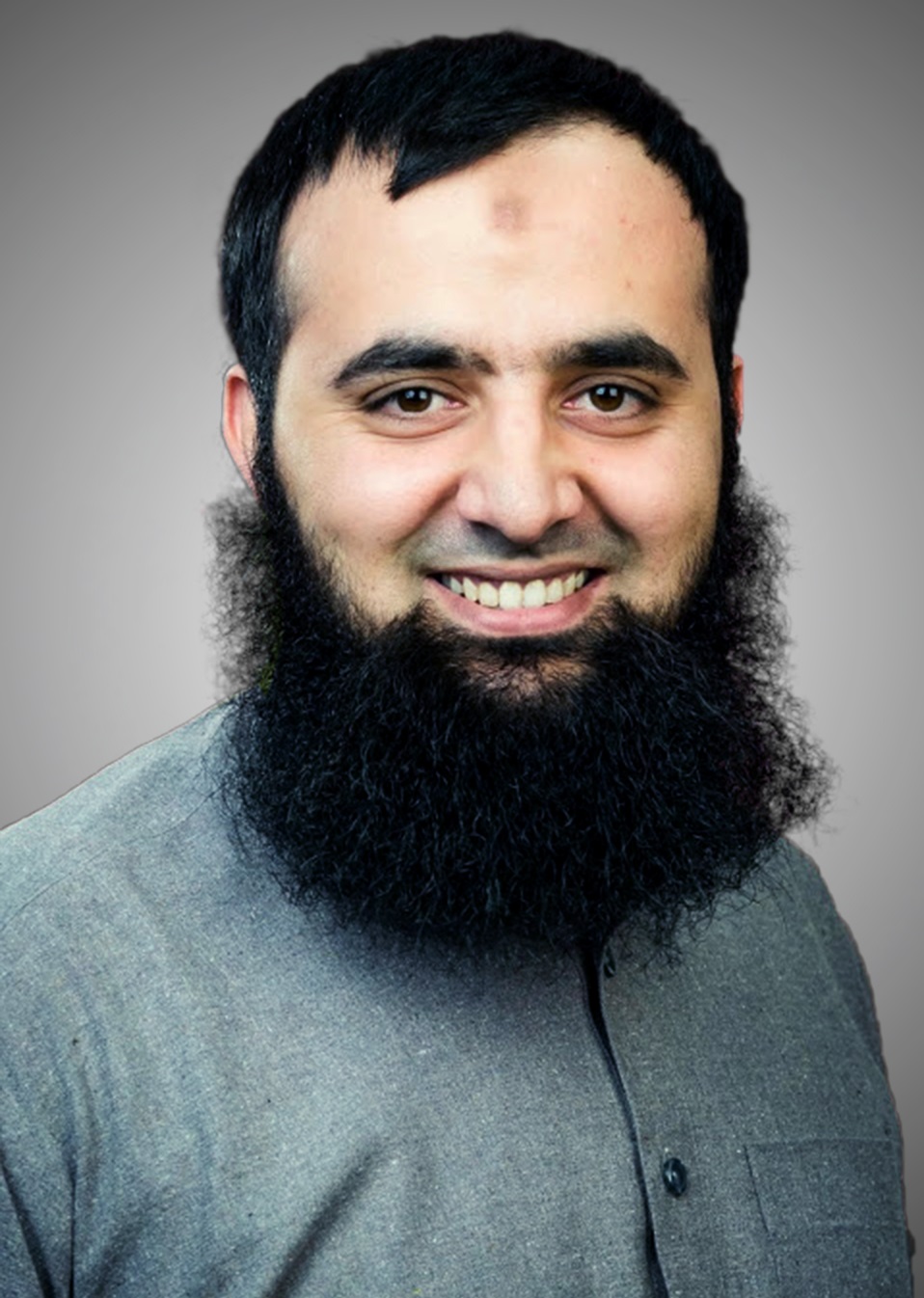}}]{Dr. Hafeez Anwar}
is Associate Professor at the National University of Computer and Emerging Scieces (FAST-NUCES) at Peshawar, KPK, 25000, Pakistan. His research interests include computer vision, deep learning and applications of computer vision. Dr. Anwar received his PhD Degree in Visual Computing from The Vienna University of Technology (TU Wien) in 2015. Contact him at hafeez.anwar@nu.edu.pk.
\end{IEEEbiography}

\end{document}